\def\year{2020}\relax
\documentclass[letterpaper]{article} 
\usepackage{aaai20}  
\usepackage{times}  
\usepackage{helvet} 
\usepackage{courier}  
\usepackage[hyphens]{url}  
\usepackage{graphicx} 
\usepackage{graphicx}  
\usepackage{amsmath}
\usepackage{amssymb}
\usepackage{graphics}
\usepackage{multirow}
\newcommand{\citet}[1]
{\citeauthor{#1}~\shortcite{#1}}
\newcommand{\citep}{\cite}

\usepackage{pifont}
\newcommand{\xmark}{\ding{55}}
\usepackage{color, colortbl}
\definecolor{Gray}{gray}{0.9}
\title{Knowing What, How and Why: A Near Complete Solution for Aspect-based Sentiment Analysis}
\author{\Large \textbf{ Haiyun Peng\textsuperscript{\rm 1}, Lu Xu\thanks{Lu Xu is under the Joint PhD Program between Alibaba and Singapore University of Technology and Design.}\textsuperscript{\rm 1,2}, Lidong Bing\textsuperscript{\rm 1}, Fei Huang\textsuperscript{\rm 1}, Wei Lu\textsuperscript{\rm 2}, Luo Si\textsuperscript{\rm 1}}\\
\textsuperscript{\rm 1} DAMO Academy, Alibaba Group \\
\textsuperscript{\rm 2} Singapore University of Technology and Design\\
{\{haiyun.p, lu.x, l.bing, f.huang, luo.si\}}@alibaba-inc.com, luwei@sutd.edu.sg 
}
\begin{document}

\maketitle

\begin{abstract}
Target-based sentiment analysis or aspect-based sentiment analysis (ABSA) refers to addressing various sentiment analysis tasks at a fine-grained level, which includes but is not limited to aspect extraction, aspect sentiment classification, and opinion extraction.  There exist many solvers of the above individual subtasks or a combination of two subtasks, and they can work together to tell a complete story, i.e. the discussed aspect, the sentiment on it, and the cause of the sentiment.  However, no previous ABSA research tried to provide a complete solution in one shot. In this paper, we introduce a new subtask under ABSA, named aspect sentiment triplet extraction (\textbf{ASTE}). Particularly, a solver of this task needs to extract triplets (What, How, Why) from the inputs, which show WHAT the targeted aspects are, HOW their sentiment polarities are and WHY they have such polarities (i.e. opinion reasons). For instance, one triplet from ``Waiters are very friendly and the pasta is simply average'' could be (`Waiters', positive, `friendly'). We propose a two-stage framework to address this task. The first stage predicts what, how and why in a unified model, and then the second stage pairs up the predicted what (how) and why from the first stage to output triplets. In the experiments, our framework has set a benchmark performance in this novel triplet extraction task. Meanwhile, it outperforms a few strong baselines adapted from state-of-the-art related methods. 
\end{abstract}

\section{Introduction}\label{sec:introduction}
\begin{figure}[!t]
        \centering
        \includegraphics[width=.95\columnwidth]{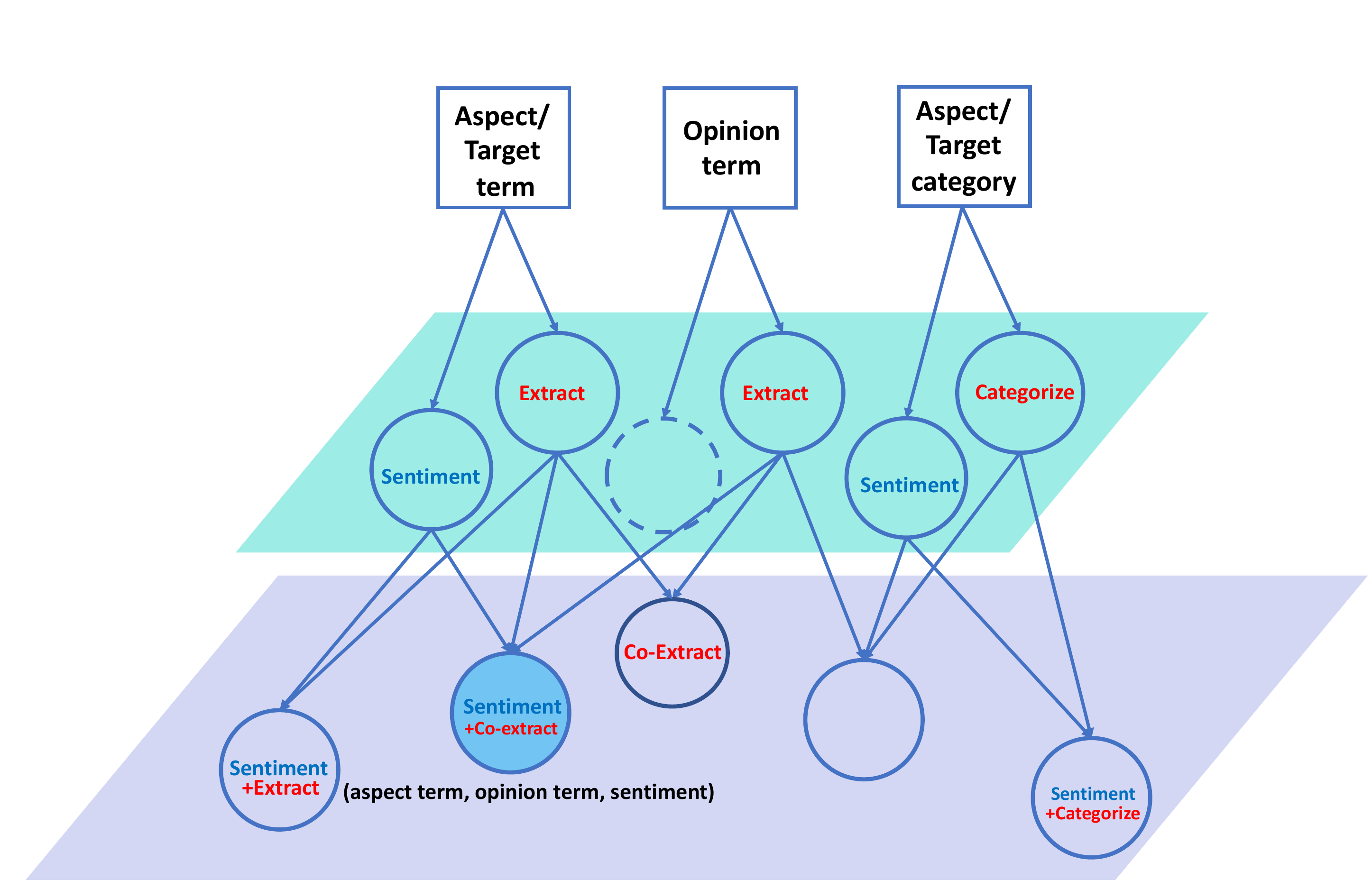}
        \caption{The road map to aspect-based sentiment analysis tasks. The bottom blue-filled circle anchors our task.}
        \label{fig:tasks}
    \end{figure}

Target-based sentiment analysis (TBSA) or aspect-based sentiment analysis (ABSA\footnote{Interchangeable with TBSA in this paper.}) refers to addressing various sentiment analysis tasks at a fine-grained level~\cite{liu2012sentiment,S14-2004}, which includes but is not limited to aspect/target term extraction (ATE), opinion term extraction (OTE), aspect/target term sentiment classification (ATC), etc. Given an example sentence such as \textit{`Waiters are very friendly and the pasta is simply average'}, the ATE is to extract `\textit{Waiters}' and `\textit{pasta}', and the ATC is to classify them to positive and negative sentiment, respectively. The OTE is to extract `\textit{friendly}' and `\textit{average}'.
Although these tasks seem to be intersecting and confusing at the first glance, they distinguish from each other black and white when fulfilling the three goals in ABSA. As shown in Fig~\ref{fig:tasks}, the top three squares represent ultimate goals for ABSA, where the aspect term represents an explicit mention of discussed target, such as `\textit{\textbf{Waiters}}' in the example. The opinion term represents the opinionated comment terms/phrases, like `\textit{\textbf{friendly}}'. The aspect category refers to certain predefined categories, such as \textbf{SERVICE} and \textbf{FOOD} in the previous example~\cite{wang2019aspect,S15-2082}.

\begin{table*}[!t]
\caption{Tagging schema and relative position index, where B denotes begin, I denotes inside, E denotes end, S denotes single and O denotes outside. The check and cross marks denote valid and invalid aspect-opinion pairs.}
\label{tab:tag-schema}
\resizebox{2.1\columnwidth}{!}{%
\begin{tabular}{c|c|ccccccccccccc}
\hline
\multicolumn{2}{c|}{Input}                                                                            & Waiters & are & friendly & and & the & fugu  & sashimi & is & out & of & the & world & . \\ \hline
\multicolumn{2}{c|}{Unified tag (aspect+sentiment)}        & S-POS   & O   & O        & O   & O   & B-POS & E-POS   & O  & O   & O  & O   & O     & O \\
\multicolumn{2}{c|}{Opinion tag}                                                                      & O       & O   & S        & O   & O   & O     & O       & O  & B   & I  & I   & E     & O \\ \hline
\multirow{2}{*}{Position index} & (Waiters,friendly)${\text{\checkmark}}$     & 2       & 0   & 2        & 0   & 0   & 0     & 0       & 0  & 0   & 0  & 0   & 0     & 0 \\
                                                                           & (fugu sashimi, friendly)${\text{\xmark}}$ & 0       & 0   & 3        & 0   & 0   & 3     & 3       & 0  & 0   & 0  & 0   & 0     & 0 \\ \hline
\end{tabular}
}
\end{table*}    

Each circle in the middle layer denotes a direct subtask to realize the goal. The `Sentiment' circle linked to aspect terms refers to ATC which attracts a heated research popularity~\cite{P14-2009,D16-1021,nguyen2015phrasernn,D16-1058,ma2017interactive,tay2017learning,ma2018targeted,N18-2043,P18-1087,P18-1088,P18-1234,P18-2092,peng2018learning,bailin-lu:2018:AAAI2018,li2019exploiting}. The `Extract' circle linked to aspect term denotes ATE, such as~\cite{J11-1002,P13-1172,P14-1030,D15-1168,yin2016unsupervised,D16-1059,wang2017coupled,P17-1036,D17-1310,li2018aspect,P18-2094}. The same also applies to other circles in the middle layer. Researchers also realized that solving these subtasks individually is insufficient so they proposed to couple two subtasks as a compound task, such as aspect term extraction and sentiment classification~\cite{li2019unified,D13-1171,D15-1073,li2017learning,he_acl2019}, aspect term and opinion term co-extraction~\cite{wang2017coupled,dai2019neural}, aspect category and sentiment classification~\cite{hu2018can}, as circles illustrated at the bottom.

Nevertheless, the above compound tasks are still not enough to get a complete picture regarding sentiment. For instance, in the previous example, knowing a positive sentiment towards aspect term `\textit{waiters}' does not give a clue of why it is positive. Only by knowing `\textit{friendly}' will people understand the cause of sentiment. \citet{fan2019target} aim to extract the opinion terms for a given target, thus the extraction can be regarded as the cause for certain sentiment on the target, through sentiment prediction is not in the scope of their paper. Note that \citet{fan2019target} assume the targets are given in advance. On the other hand, the co-extraction methods fail to tackle pairing of multiple aspects and opinion expressions in a single sentence~\cite{wang2017coupled,dai2019neural}. 
\citet{li2019unified} couple the tasks of aspect extraction and sentiment classification with the unified tags (e.g. ``B-POS'' standing for the beginning of a positive aspect) but they do not extract the opinion terms for the extracted aspects, leaving blank the sentiment cause. So did the modular architectures presented by~\citet{zhang2019sentiment}. In summary, no previous ABSA research try to handle such a requirement in one shot, namely knowing \textbf{\textit{What}} target is being discussed (e.g. `waiters'), \textbf{\textit{How}} is the sentiment  (e.g. `positive') and \textbf{\textit{Why}} is this sentiment (e.g. `friendly'). 
Moreover, the mutual influence among the three questions lacks study either.  To this end, we introduce an aspect sentiment triplet extraction task (\textbf{ASTE}), shown in the blue-filled circle at the bottom in Fig~\ref{fig:tasks}.

We propose a two-stage framework to address this task. In the first stage, we aim to extract potential aspect terms, together with their sentiment, and extract potential opinion terms. 
The task is formulated as a labeling problem with two label sequences~\cite{D13-1171,wang2017coupled}. Specifically, we couple a unified tagging system by following \cite{li2019unified} for aspect extraction and sentiment classification, and a BIO-like tagging system for opinion extraction, as shown in Table~\ref{tab:tag-schema}. For the unified tagging system, it builds on top of two stacked Bidirectional Long Short Term Memory (BLSTM) networks. The upper one produces the aspect term and sentiment tagging results based on the unified tagging schema. The lower performs an auxiliary prediction of aspect boundaries with the aim for guiding the upper BLSTM. Gate mechanism is explicitly designed to maintain the sentiment consistency within each multi-word aspect. For the opinion term tagging system, it builds on top of a BLSTM layer and a Graph Convolutional Network (GCN) to make full use of semantic and syntactic information in a sentence.
According to the task definition \cite{S14-2004,S15-2082,S16-1002,li2018aspect,li2019unified}, for a term/phrase being regarded as an aspect, it should co-occur with some ``opinion terms'' that indicate a sentiment polarity on it. 
Therefore, aspect information is beneficial to extracting opinion terms, as already demonstrated in~\cite{zhang2017semi,wang2017coupled,dai2019neural}. 
We specifically design a target-guiding module to transfer aspect information for opinion term extraction. 

After the first stage, we have obtained a bunch of aspects with sentiment polarities and a bunch of opinion expressions. In the second stage, the goal is to pair up aspects with the corresponding opinion expressions. As we observed, for sentences with multiple aspects and opinions, word distance is very indicative for correctly pairing up an aspect and its opinion as shown in the bottom section of Table \ref{tab:tag-schema}. 
Thus, we design distance embeddings to capture the distance between aspects and opinion expressions that are predicted from the stage one. 
With a BLSTM encoder, we encode sentence-level contexts into aspect and opinion terms for the final classification of candidate pairs.
In the experiments, our framework has set a benchmark performance in this novel sentiment triplet extraction task. Meanwhile, our framework outperforms the state-of-the-art methods (with modification to fit in our task) and the strongest sequence taggers on several benchmark datasets. We also conduct extensive ablation tests to validate the rationality of our framework design.

\section{Proposed Framework}\label{sec:proposedframwork}

\subsection{Problem Formulation}
For a given input sentence $\mathrm{X}=\{x_1,\dots,x_T\}$ with length $T$, the \textbf{ASTE} task is to extract sentiment triplets (What, How, Why), consisting of the aspects/targets (i.e. `What'), the sentiment polarity on them (i.e. `How'), and the opinions causing such a sentiment (i.e. `Why').~\footnote{Note that the opinion expressions should be paired with the targets/aspects it modifies in a many-to-many setting.}
Here, we formulate the task in two stages.
In the stage one, the task includes two sequence labeling (SL) subtasks, the unified tag SL and the opinion tag SL. The unified tag schema is $\mathcal{Y}^{\mathcal{TS}} = \{\texttt{B-POS}, 
\texttt{I-POS},\texttt{E-POS},\texttt{S-POS}, \texttt{B-NEG}, \texttt{I-NEG},\\ \texttt{E-NEG}, \texttt{S-NEG}, \texttt{B-NEU},\texttt{I-NEU},\texttt{E-NEU}, \texttt{S-NEU} \} \cup \{\texttt{O}\}$, which locates aspects and labels their sentiment.
The opinion tag schema is $\mathcal{Y}^{\mathcal{OPT}} = \{\texttt{B},\texttt{I},\texttt{E},\texttt{S} \} \cup \{\texttt{O}\}$. The unified SL predicts a tag sequence $\mathrm{ Y}^{\mathcal{TS}}=\{y^{\mathcal{TS}}_1,\dots,y^{\mathcal{TS}}_T\}$, where $y^{\mathcal{TS}}_i \in \mathcal{Y}^{\mathcal{TS}}$, while the opinion SL predicts a sequence $\mathrm{ Y}^{\mathcal{OPT}}=\{y^{\mathcal{OPT}}_1,\dots,y^{\mathcal{OPT}}_T\}$, where $y^{\mathcal{OPT}}_i \in \mathcal{Y}^{\mathcal{OPT}}$.

    \begin{figure}[!t]
        \centering
        \includegraphics[width=.9\columnwidth]{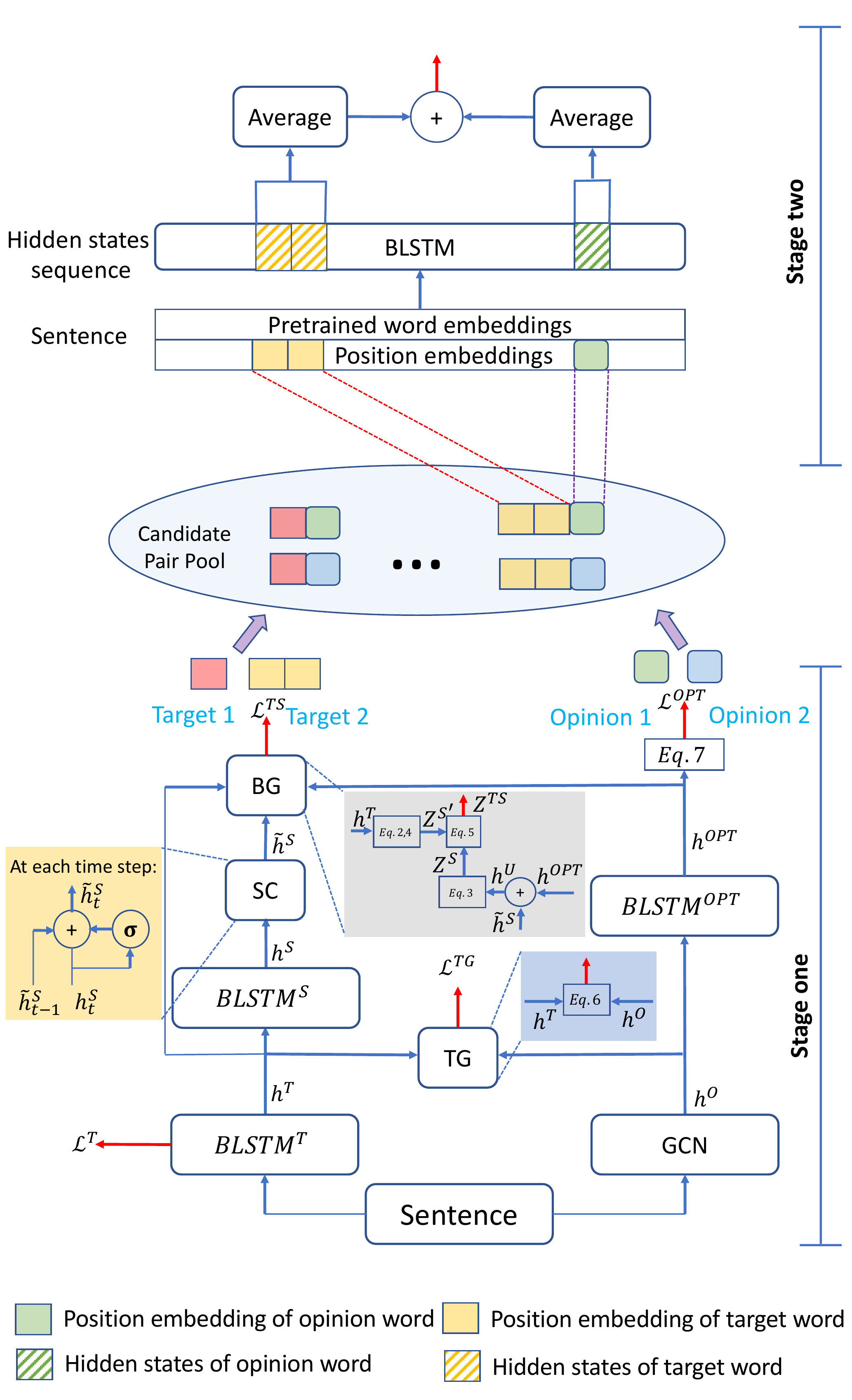}
        \caption{The framework of our proposed two stage model.}
        \label{fig:framework}
    \end{figure}

In the stage two, given the sets of aspects $\{T_1, T_2 ,..., T_n\}$ and opinion expressions $\{O_1, O_2 ,..., O_m\}$ labeled from the same sentence in the stage one, where there are $n$ aspects \footnote{Each aspect could contain one or multiple terms. The same also applies to the opinion expression. In our model setting, there should be at least one aspect and one opinion expression.} and $m$ opinion expressions, a candidate pair pool is constructed by coupling the elements from the two sets as $\{(T_1,O_1), (T_1,O_2), ... , (T_n,O_m)\}$. The goal of this stage is to identify the legitimate ones from the candidate pool, and outputs them as the final results. Note that the `How' is embedded in $T_i$ with the unified tags.

\subsection{Model Overview}
Fig.~\ref{fig:framework} shows the overview of our two-stage framework. 
Recall that the stage one predicts two kinds of labels, i.e. $\mathcal{Y}^{\mathcal{TS}}$ and $\mathcal{Y}^{\mathcal{OPT}}$. 
For predicting $\mathcal{Y}^{\mathcal{TS}}$, the left half of stage one model resembles the state-of-the-art work \cite{li2019unified} for unified tag schema,
and adapts one of its original component as a shared one (i.e. TG) with the right part for predicting $\mathcal{Y}^{\mathcal{OPT}}$. 
Specifically, the left side contains two stacked BLSTM. The lower one BLSTM$^T$ performs an auxiliary prediction of target boundaries (i.e. BIO) for producing signals for the upper BLSTM, the boundary guidance (BG), and the target guidance (TG). The hidden states from the upper BLSTM$^S$ are first manipulated by the sentiment consistency (SC), and then used as the major signal to predict the unified tags by the BG component, which also transforms the pure target boundary tag prediction to guide unified tag prediction. Our design distinguishes from \citet{li2019unified} in the specific injection of opinion information for predicting unified tag (i.e. $h^{OPT}$ is used by BG).

The right part of the stage one is for the opinion term prediction, i.e. $\mathcal{Y}^{\mathcal{OPT}}$. The sentence is fed into a GCN to learn the mutual influence of target and opinion terms via dependency relations\footnote{https://spacy.io/}.
Afterwards, this signal will be sent to two different modules, TG and $BLSTM^{OPT}$. The TG component in the middle is the concatenation of pure target boundary information and the GCN output, which leverages the target information for opinion term extraction. Unlike~\citet{li2019unified} whose opinion information is weak supervision from sentiment lexicon lookup, our design specifically constructs a component sharing both target and opinion information. This component is strongly supervised by opinion term extraction, therefore, both $BLSTM^{T}$ and GCN can benefit from its backpropagation. Meanwhile, the output from $BLSTM^{OPT}$ will carry the sentence context on top of the GCN output. It will be sent for opinion term extraction, as well as for guiding unified tag prediction.

The stage two model firstly uses the aspects and opinion expressions predicted from stage one to generate all possible pairs in each sentence. Based on the distance between target and opinion expression in each pair, a position embedding is applied for each target and opinion terms. Non-target/non-opinion term will have the same position embedding, which is zero in our experiments. After a BLSTM encoder, the hidden states from the aspect and opinion expressions will be concatenated for binary classification.

\subsection{Stage One} 

\subsubsection{Unified aspect boundary and sentiment labeling.}

As demonstrated by \citet{li2019unified}, a target boundary tag is beneficial to unified tag prediction. We implement a similar structure for unified aspect and sentiment tag labeling. 
In order to learn the target boundary labeling, we employ a $\text{BLSTM}^{\mathcal{T}}$ layer on top of sentence word embeddings. 
The sequence output of this $\text{BLSTM}^{\mathcal{T}}$,
$h^{\mathcal{T}} = [\overrightarrow{\text{LSTM}}^{\mathcal{T}}(x); \overleftarrow{\text{LSTM}}^{\mathcal{T}}(x)]$, 
will be fed into a softmax classifier to predict the target sequence tag without sentiment, whose tag set
$\mathcal{Y}^{\mathcal{T}}$ is \{\texttt{B}, \texttt{I}, \texttt{E}, \texttt{S}, \texttt{O}\}. With this supervision learning, 
$h^{\mathcal{T}}$ is expected to carry target boundary information. Thus we input it to the second $\text{BLSTM}^{\mathcal{S}}$ layer to accumulate sentiment information. 
Specifically, the sequence output of this $\text{BLSTM}^{\mathcal{S}}$ is $h^{\mathcal{S}} = [\overrightarrow{\text{LSTM}}^{\mathcal{S}}(x); \overleftarrow{\text{LSTM}}^{\mathcal{S}}(x)]$. The expected tag set for each time step in this sequence is $\mathcal{Y}^{\mathcal{S}} = \{\texttt{B-POS},\texttt{I-POS},\texttt{E-POS},\texttt{S-POS}, \texttt{B-NEG}, \texttt{I-NEG}, \\ \texttt{E-NEG},\texttt{S-NEG}, \texttt{B-NEU},\texttt{I-NEU},\texttt{E-NEU}, \texttt{S-NEU} \}$, which appends each tag in $\mathcal{Y}^{\mathcal{T}}$ with three sentiment polarities.

As illustrated in Table~\ref{tab:tag-schema}, some aspects may contain more than one term. In our task formulation, however, we predict unified sequence tag term by term. It is possible, although contradictory, to have `\textit{fugu}' labeled as positive but `\textit{sashimi}' labeled as negative. To avoid such situation, a Sentiment Consistency (SC) module~\cite{li2019unified} was designed with a gate mechanism:
\begin{equation}
\begin{split}
    g_t &= \sigma(\mathrm{\bf W}^g h^{\mathcal{S}}_t + \mathrm{\bf b}^g)\\
    \tilde{h}^{\mathcal{S}}_t &= g_t \odot h^{\mathcal{S}}_t + (1-g_t) \odot \tilde{h}^{\mathcal{S}}_{t-1}
    \end{split}
\end{equation}
where $\mathrm{\bf W}^g$ and $\mathrm{\bf b}^g$ are model parameters of the SC module, and $\odot$ is the element-wise multiplication. $\sigma$ is a sigmoid function. With this gate mechanism, current time step prediction will also inherit features from the previous time step, reducing the risk of drastic sentiment label change.

Given an aspect boundary tag, it can only be transformed into one of its three legitimate sentiment-appended unified tags. The Boundary Guidance (BG) module consolidates this observation into a constraint matrix transformation $\text{\bf W}^{tr} \in \mathbb{R}^{|\mathcal{Y}^{\mathcal{T}}| \times |\mathcal{Y}^{\mathcal{S}}|}$. This is a probability transformation matrix in which $\text{\bf W}^{tr}_{i,j}$ indicates the probability of tag $\mathcal{Y}^{\mathcal{T}_{i}}$ transforming to tag $\mathcal{Y}^{\mathcal{S}_{j}}$. For instance, if $\mathcal{Y}^{\mathcal{T}_{i}}$ is $\texttt{I}$ and $\mathcal{Y}^{\mathcal{S}_{j}}$ is $\texttt{B-POS}$, then the $\text{\bf W}^{tr}_{\texttt{I},\texttt{B-POS}}$ will be zero because $\texttt{I}$ cannot transform to $\texttt{B-POS}$. With this transformation matrix, the aspect boundary probability distribution can now be transformed into unified probability distribution as:
\begin{equation}
\begin{gathered}
    z^{\mathcal{T}}_t = {\bf p}(y^{\mathcal{T}}_t|x_t) = \mathrm{Softmax}(\mathrm{\bf W}^{\mathcal{T}} h^{\mathcal{T}}_t)\\
    z^{\mathcal{S}^{'}}_t = (\text{\bf W}^{tr})^{\top}z^{\mathcal{T}}_t
\end{gathered}
\end{equation} 
$\text{\bf W}^{\mathcal{T}}$ is the model parameter and $z^{\mathcal{S}^{'}}_t$ is the obtained unified tag probability distribution.

Up to this moment, for the unified tag prediction, we have not directly utilized the opinion term information, which should apparently affect the detection of aspect. To this end, we integrate the opinion term information ${h}^{\mathcal{OPT}}$ (which will be introduced in the next subsection) with $\tilde{h}^{\mathcal{S}}$  by concatenating them together to form a reinforced representation ${h}^{\mathcal{U}}$ for unified tag prediction with a softmax classifier:
\begin{equation}\label{eq3}
    z^{\mathcal{S}}_t = {\bf p}(y^{\mathcal{S}}_t|x_t) = \mathrm{Softmax}(\mathrm{\bf W}^{\mathcal{S}} h^{\mathcal{U}}_t)
\end{equation}
where the $\mathcal{Y}^{\mathcal{S}}$ is the probability distribution and $\text{\bf W}^{\mathcal{S}}$ is the model parameter. $z^{\mathcal{S}}_t$ is the obtained unified tag probability distribution. Note that the integration of ${h}^{\mathcal{OPT}}$ for unified tag prediction was not used in \citet{li2019unified}.

Next, we design a fusion mechanism to merge this reinforced unified tag probability with the previous transformed unified tag probability. We calculate a fusion weight score $\alpha_t \in \mathbb{R}$ with the concentration score $c_t$ from the target boundary tagger, defined as below:
\begin{equation}
\begin{split}
    c_t &= (z^{\mathcal{T}}_t)^{\top} z^{\mathcal{T}}_t \\
    \alpha_t &= \epsilon c_t
\end{split}
\end{equation}

where the concentration score $c_t$, with a maximum value of 1, represents how confident the target boundary tagger predicts. The higher the score, the more confident is the target boundary tagger. The hyper-parameter $\epsilon$ (we empirically set as 0.5) controls the proportional weight that transformed unified tag probability contributes in the final decision. Then the final fused score between transformed and reinforced unified tag probability is given as:

\begin{equation}
    {z}^{\mathcal{TS}}_t = \alpha_t z^{\mathcal{S}^{'}}_t + (1-\alpha_t) z^{\mathcal{S}}_t.
\end{equation}

\subsubsection{Opinion term extraction.}
Previous studies~\cite{wang2017coupled,dai2019neural} suggest that aspect extraction and opinion extraction are mutually beneficial. We also observe that aspects are usually co-occur with opinion terms and especially so on our datasets (see Table~\ref{tab:dataset}). This drives us to utilize the target information to guide opinion term extraction.
Particularly, we feed the sentence embedding to a GCN module to learn the mutual dependency relations between different words. The adjacency matrix for GCN is constructed based on the dependency parsing of the sentence, namely $\text{\bf W}^{GCN} \in \mathbb{R}^{^{{\mathcal{|L|}} \times {\mathcal{|L|}}}}$, where $\mathcal{L}$ is the length of the sentence. If the $i$th word has dependency relation with the $j$th word, $\text{\bf W}^{GCN}_{i,j}$ and $\text{\bf W}^{GCN}_{j,i}$ will both have value 1, otherwise, value 0. This operation is designed to capture the relation between aspects and opinion terms, as they are constructed as syntactic modifying pairs.

To utilize the target information for opinion term extraction, we design an auxiliary task to integrate the target boundary information with the output from GCN with a Target Guidance (\textbf{TG}) module. If a sentence contains an aspect-opinion pair, the opinion expression should modify its aspect following syntactic rules. Thus, given a target signal from $BLSTM^{T}$, it is intuitive to use it to guide opinion term extraction. We have tried various implementations of TG, and in the end a simple concatenation achieved the best performance. The concatenation will be fed into a softmax classifier for opinion tag classification in the tag space of $\mathcal{Y}^{\mathcal{TG}} = \{\texttt{B},\texttt{I},\texttt{E},\texttt{S} \} \cup \{\texttt{O}\}$:
\begin{equation}
    z^{\mathcal{TG}}_t = {\bf p}(y^{\mathcal{OPT}}_t|x_t) = \mathrm{Softmax}(\mathrm{\bf W}^{\mathcal{TG}} [h^{\mathcal{T}}_t;h^{\mathcal{O}}_t]).
\end{equation}

Next, the sequence of hidden states from GCN ($h^{O}$) is sent to a $\text{BLSTM}^{\mathcal{OPT}}$ for sequence learning, namely to encode the contextual information within the sentence, and the output, $h^{OPT}$, will be sent to both the \textbf{BG} component to assist unified tag prediction (Eq.\ref{eq3}) and a softmax classifier to predict opinion prediction:
\begin{equation}
    z^{\mathcal{OPT}}_t = {\bf p}(y^{\mathcal{OPT}}_t|x_t) = \mathrm{Softmax}(\mathrm{\bf W}^{\mathcal{OPT}} h^{\mathcal{OPT}}_t).
\end{equation}

\subsubsection{Stage one training.}
Stage one is trained with stochastic gradient descent optimizer. The loss of each output signal is computed using crossentropy as:
\begin{equation}
    \mathcal{L}^{\mathcal{I}} = -\frac{1}{T} \sum^{T}_{t=1} \mathbb{I}(y^{\mathcal{I},g}_t) \circ \log(z^{\mathcal{I}}_t)
\end{equation}
where $\mathcal{I}$ is the symbol of task indicator and its possible values are $\mathcal{T}$, $\mathcal{TS}$, $\mathcal{TG}$ and $\mathcal{OPT}$. $\mathbb{I}(y)$ represents the one-hot vector with the $y$-th component being 1 and $y^{\mathcal{I},g}_t$ is the gold standard tag for the task $\mathcal{I}$ at the time step $t$. The total training objective of stage one is to minimize the sum of individual loss from each output signal, $\mathcal{J}(\theta)$:
\begin{equation}
    \mathcal{J}(\theta) = \mathcal{L}^{\mathcal{T}} + \mathcal{L}^{\mathcal{TS}} + \mathcal{L}^{\mathcal{TG}}+ \mathcal{L}^{\mathcal{OPT}}.
\end{equation}

\subsection{Stage Two}
After stage one, for each sentence, we output two sets of text segments, i.e., aspect terms and opinion expressions, denoted as $\{T_{1}, T_{2}, ..., T_{n}\}$ and $\{O_{1}, O_{2}, ..., O_{m}\}$ respectively, where there are $n$ aspects and $m$ opinion expressions. Then, we generate a candidate pair pool as $\{(T_1,O_1), (T_1,O_2), ... , (T_n,O_m)\}$ by enumerating all possible aspect-opinion pairs. Stage two is to classify whether each of these pair is valid or not.

\subsubsection{Position embeddings.}
In order to utilize the position relation between an aspect and an opinion expression, we calculate the word-length distance between the center of the aspect and that of the opinion expression by counting how many words appear in the middle. The absolute distance will be treated as relative position information that encodes the position relation between them. For the ease of training, we create position embeddings by treating the distance as position index for aspects and opinions, and zero to non-aspect and non-opinion words. For instance, position indexes of a true pair (`\textit{Waiter}',`\textit{friendly}') and a fake pair (`\textit{fugu sashimi}',`\textit{friendly}') are shown in Table~\ref{tab:tag-schema}.

\subsubsection{Pair encoder and classification.}
As shown in Fig.~\ref{fig:framework}, we concatenate the pretrained GloVe word embeddings~\cite{D14-1162} with our position embeddings to form word representation. The position embedding is randomly initialized and kept trainable in the training step. We then feed the sentence to a BLSTM layer to encode sentence contextual information into aspects and opinion expressions. Based on the sentence term index, we average the hidden states output from BLSTM for both aspect and opinion expression respectively as their features. Next, we concatenate the two features and send it to softmax layer for binary classification. 
For the training of classifier, we used the gold pairs annotated in the training set of our experimental datasets. During testing stage, we freeze the classifier parameters tuned against the validation sets, and directly test on the pairs generated in the candidate pool.

\section{Experiments}\label{sec:experiments}
\subsection{Dataset}
Our datasets \footnote{\url{https://github.com/xuuuluuu/SemEval-Triplet-data}} originate from SemEval Challenges~\cite{S14-2004,S15-2082,S16-1002}. The annotation (opinion label) is derived from \cite{fan2019target}, where they already annotated opinion terms. In addition, we merge samples that are of the same sentence but have different annotations on targets and opinions. Each sample includes the original sentence, a sequence with unified aspect/target tags and a sequence with opinion tags. Since each sentence might have more than one aspect/targets and opinions, we pair up individual aspects/targets and their opinions. Below is an example:\vspace{2mm}\\
\textit{
The best thing about this laptop is the price along with some of the newer features . \vspace{1mm}\\
The=O best=O thing=O about=O this=O laptop=O is=O the=O price=T-POS along=O with=O some=O of=O the=O newer=O features=TT-POS .=O \vspace{1mm}\\
The=O best=S thing=O about=O this=O laptop=O is=O the=O price=O along=O with=O some=O of=O the=O newer=SS features=O .=O\vspace{2mm}\\}
The example consists of two  target and opinion pairs, the first pair is `price' and `best', the second pair is `feature' and `newer'. Note that `TT-POS' is only used for indicating the pairing relation with `SS', for model training, the used tags are `T-POS' and `S'.
We also correct a small number of samples whose targets and opinions are overlapped. The validation set is randomly selected 20\% of data from training set. Table~\ref{tab:dataset} shows the detailed statistics.
\begin{table}[t]
\caption{Dataset. (\#s and \#p denote number of sentences and target-opinion pairs, respectively.)}
\label{tab:dataset}
\resizebox{\columnwidth}{!}{%

\begin{tabular}{l|l|l|l|l|l|l|l|l}
\hline
\multicolumn{1}{c|}{\multirow{2}{*}{Dataset}} & \multicolumn{2}{c|}{14res} & \multicolumn{2}{c|}{14lap} & \multicolumn{2}{c|}{15res} & \multicolumn{2}{c}{16res} \\ \cline{2-9} 
\multicolumn{1}{c|}{}                         & \#s    & \# p   & \#s    & \#p    & \#s    & \#p    & \#s    & \#p    \\ \hline
train                                          & 1300          & 2145       & 920           & 1265       & 593           & 923        & 842           & 1289       \\ \hline
valid                                          & 323           & 524        & 228           & 337        & 148           & 238        & 210           & 316        \\ \hline
test                                           & 496           & 862        & 339           & 490        & 318           & 455        & 320           & 465        \\ \hline
\end{tabular}
}
\end{table}

\begin{table*}[t]
\caption{Stage one results of aspect extraction and sentiment classification.  (All models were trained in the unified tag setting.) }
\label{tab:stage1_ts}
\centering
\begin{small}
\begin{tabular}{l|ccc|ccc|ccc|ccc}
\hline
               & \multicolumn{3}{c|}{14res}                       & \multicolumn{3}{c|}{14lap}                       & \multicolumn{3}{c|}{15res}                       & \multicolumn{3}{c}{16res}                        \\ \cline{2-13} 
                                   & P              & R              & F              & P              & R              & F              & P              & R              & F              & P              & R              & F              \\ \hline
RINANTE                            & 48.97          & 47.36          & 48.15          & 41.20          & 33.20          & 36.70          & 46.20          & 37.40          & 41.30          & 49.40          & 36.70          & 42.10          \\
CMLA                               & 67.80              & 73.69              & 70.62              & 54.70              & 59.20              & 56.90              & 49.90              & 58.00              & 53.60              & 58.90              & 63.60              & 61.20              \\
Li-unified                         & 74.43          & 69.26          & 71.75          & 68.01 & 56.72          & 61.86          & 61.39          & 67.99 & 64.52          & 66.88          & 71.40          & 69.06          \\
Li-unified-R                       & 73.15          & 74.44          & 73.79          & 66.28          & 60.71          & \textbf{63.38} & 64.95          & 64.95          & 64.95          & 66.33          & 74.55 & 70.20          \\ \hline
Our--$\text{BLSTM}^{\mathcal{OPT}}$ & 70.00          & 74.20          & 72.04 & 65.99          & 54.62          & 59.77          & 63.41          & 65.19 & 64.29          & 69.74          & 71.62          & 70.67          \\
Our--TG                             & 74.41 & 73.97          & \textbf{74.19}          & 64.35          & 60.29          & 62.26          & 59.28          & 61.92          & 60.57          & 64.57          & 66.89          & 65.71          \\
Our--T       				           & 69.42 & 72.2          & 70.79          & 64.14          & 60.63          & 62.34         & 62.28         & 66.35         & 64.25         & 62.65          & 71.4         & 66.74         \\
Our                                & 76.60          & 67.84 & 71.95 & 63.15          & 61.55 & 62.34 & 67.65 & 64.02          & \textbf{65.79} & 71.18 & 72.30  & \textbf{71.73} \\ \hline\hline
\multicolumn{13}{c}{The two rows below are results of aspect extraction only, without evaluating the correctness of sentiment polarity.} \\\hline
\rowcolor{Gray}
RINANTE                            & 75.89          & 70.34          & 73.00          & 70.80          & 52.80          & 60.50          & 72.64          & 51.68          & 60.39          & 67.10          & 55.20          & 60.60          \\
\rowcolor{Gray}
CMLA                               & 84.21		          & 89.83          & 86.93          & 71.50		          & 82.20          & 76.40          & 75.10		          & 89.30          & 81.50          & 72.00		          & 87.60          & 79.00          \\ \hline
\end{tabular}
\end{small}
\end{table*}


\begin{table*}[t]
\caption{Stage one results of opinion term extraction.}
\label{tab:stage1_opt}
\begin{small}
\centering
\begin{tabular}{l|ccc|ccc|ccc|ccc}
\hline
 & \multicolumn{3}{c|}{14res}                       & \multicolumn{3}{c|}{14lap}                       & \multicolumn{3}{c|}{15res}              & \multicolumn{3}{c}{16res}                       \\ \cline{2-13} 
                  & P              & R              & F              & P              & R              & F              & P              & R              & F     & P              & R              & F              \\ \hline
Distance rule     & 58.39          & 43.59          & 49.92          & 50.13          & 33.86          & 40.42          & 54.12          & 39.96          & 45.97 & 61.90          & 44.57          & 51.83          \\
Dependency rule   & 64.57          & 52.72          & 58.04          & 45.09          & 31.57          & 37.14          & 65.49          & 48.88          & 55.98 & 76.03          & 56.19          & 64.62          \\
RINANTE           & 81.06          & 72.05          & 76.29          & 78.20          & 62.70          & 69.60          & 77.40          & 57.00          & 65.70 & 75.00          & 42.40          & 54.10          \\
CMLA              & 69.47          & 74.53          & 71.91          & 51.80          & 65.30          & 57.70          & 60.80          & 65.30          & 62.90 & 74.50          & 69.00          & 71.70          \\
IOG               & 82.85 & 77.38          & 80.02          & 73.24          & 69.63          & 71.35          & 76.06          & 70.71          & 73.25 & 85.25 & 78.51          & 81.69          \\
Li-unified-R & 81.20          & 83.18 & 82.13         & 76.62          & 74.90 & 75.70          & 79.18 & 75.88          & 77.44 & 79.84          & 86.88 & 83.16          \\\hline
Our--$\text{BLSTM}^{\mathcal{OPT}}$     & 80.41          & 86.19          & 83.15          & 78.06          & 68.98          & 73.19          & 74.29          & 80.48 & 77.21 & 82.12          & 84.95          & 83.46          \\
Our--TG            & 81.77 & 84.80          & \textbf{83.21} & 76.87          & 75.31 & \textbf{76.03} & 75.98          & 76.32          & 76.10 & 82.33 & 85.16          & 83.67          \\
Our--T     					       & 80.61 & 85.38         & 82.88 & 76.69         & 73.88 & 75.21 & 78.13          & 75.22         & 76.60 & 77.14 & 87.10         & 81.77         \\
Our               & 84.72         & 80.39 & 82.45 & 78.22 & 71.84          & 74.84          & 78.07 & 78.07 & \textbf{78.02} & 81.09          & 86.67 & \textbf{83.73} \\ \hline
\end{tabular}

\end{small}
\end{table*}

\subsection{Experimental Setting}

Our framework is evaluated on a two-stage setting due to our framework design. Since the output of our stage one contains both aspects and opinion terms, we compared with other aspect and opinion co-extraction methods in the first stage. The compared methods are as follows. \textbf{RINANTE} \cite{dai2019neural}: It is an aspect and opinion co-extraction method that mines aspect and opinion term extraction rules based on the dependency relations of words in a sentence. \textbf{CMLA} \cite{wang2017coupled}: A co-extraction model that leverages attention mechanism to utilize the direct and direction dependency relations. Note that both CMLA and RINANTE use BIO tags for aspect and opinion extraction. For comparison, we train them with unified tags for aspect extraction, and BIO tags for opinion extraction. 
     \textbf{IOG}~\cite{fan2019target}: A top performing opinion term extraction method with an Inward-Outward LSTM.
     \textbf{Li-unified}: \cite{li2019unified} The state-of-the-art unified model for aspect extraction and sentiment classification. It also serves as a base model in our design and its results are compared on aspect extraction and sentiment classification. Note that it does not conduct opinion extraction.
     \textbf{Li-unified-R}: A modified model variant of Li-unified by us, which adapts their original OE component for opinion extraction.
     \textbf{Our--$\text{BLSTM}^{\mathcal{OPT}}$}: The first variant of our model that removes the $\text{BLSTM}^{\mathcal{OPT}}$ component. Thus, it may fail to consider sentence contextual information for opinion term extraction.
     \textbf{Our--TG}: The second variant of our model that removes the TG component, which does not have the mutual information exchange between aspect extraction and opinion extraction.
     \textbf{Our--T}: The third variant that eliminates the loss $\mathcal{L}^{\mathcal{T}}$ from the training.

For the stage two evaluation, we cannot find a baseline to compare under identical settings. Thus we stack our stage two model directly on the best performed stage one baselines to construct different pipeline models. In addition to evaluating the triplets (eg. (\textit{Waiter}-\textit{friendly}-POS))\footnote{We switch TS-OPT pairs to target-opinion-sentiment triplets.}, we also evaluate the performances on the pairs (eg. (\textit{Waiter}-\textit{friendly})).

The implementations all use GloVe~\cite{D14-1162} embeddings of 300 dimension and remove domain embeddings for a fair comparison. We train up to 40 epochs with SGD optimizer with an initial learning rate 0.1 and decay rate at 0.001. Dropout rate of 0.5 is applied on the ultimate features before prediction. We report testing results of the epoch that has the best validation performance. 

\begin{table*}[t]
\caption{Stage two results in both pair and triplet setting. (+ denotes cascading our stage two module.) }
\label{tab:stage2}
\centering
\begin{small}

\begin{tabular}{c|l|ccc|ccc|ccc|ccc}
\hline
\multicolumn{2}{c|}{}                    & \multicolumn{3}{c|}{14res}     & \multicolumn{3}{c|}{14lap}     & \multicolumn{3}{c|}{15res}     & \multicolumn{3}{c}{16res}     \\ \cline{3-14} 
\multicolumn{2}{c|}{}                    & P     & R     & F              & P     & R     & F              & P     & R     & F              & P     & R     & F              \\ \hline
\multicolumn{2}{l|}{Classifier F1}       & \multicolumn{3}{c|}{97.59}     & \multicolumn{3}{c|}{94.36}     & \multicolumn{3}{c|}{99.61}     & \multicolumn{3}{c}{97.91}     \\ \hline
\multirow{4}{*}{Pair}    & RINANTE+      & 42.32 & 51.08 & 46.29          & 34.40 & 26.20 & 29.70          & 37.10 & 33.90 & 35.40          & 35.70 & 27.00 & 30.70          \\ \cline{2-2}
                         & CMLA+         & 45.17 & 53.42 & 48.95          & 42.10 & 46.30 & 44.10          & 42.70 & 46.70 & 44.60          & 52.50 & 47.90 & 50.00          \\ \cline{2-2}

                         & Li-unified-R+ & 44.37 & 73.67 & 55.34 & 52.29 & 52.94 & 52.56          & 52.75 & 61.75 & \textbf{56.85} & 46.11 & 64.55 & 53.75          \\ \cline{2-2}

                         & Our           & 47.76 & 68.10 & \textbf{56.10}          & 50.00 & 58.47 & \textbf{53.85} & 49.22 & 65.70 & 56.23          & 52.35 & 70.50 & \textbf{60.04} \\ \hline
\multirow{4}{*}{Triplet} & RINANTE+      & 31.07 & 37.63 & 34.03          & 23.10 & 17.60 & 20.00          & 29.40 & 26.90 & 28.00          & 27.10 & 20.50 & 23.30          \\ \cline{2-2}
                         & CMLA+         & 40.11 & 46.63 & 43.12          & 31.40 & 34.60 & 32.90          & 34.40 & 37.60 & 35.90          & 43.60 & 39.80 & 41.60          \\ \cline{2-2}
                         & Li-unified-R+ & 41.44 & 68.79 & 51.68 & 42.25 & 42.78 & 42.47          & 43.34 & 50.73 & 46.69          & 38.19 & 53.47 & 44.51          \\ \cline{2-2}
                         & Our           & 44.18 & 62.99 & \textbf{51.89}          & 40.40 & 47.24 & \textbf{43.50} & 40.97 & 54.68 & \textbf{46.79} & 46.76 & 62.97 & \textbf{53.62} \\ \hline
\end{tabular}

\end{small}

\end{table*}

\subsection{Results and Analysis}
\subsubsection{Stage one.}
Table~\ref{tab:stage1_ts} presents the unified performance of the stage one  for aspect extraction and sentiment classification. Our  model outperforms existing strong baselines (i.e. RINANTE, CMLA, and Li-Unified) on all datasets, especially compared with the Li-unified model which is the state-of-the-art in the unified task. Interestingly, the baseline Li-unified-R, derived from Li-unified, performs very competitive, i.e. better than Li-unified on all datasets. It shows that given the ground-truth label of opinion words, explicitly modeling opinion extraction can help upgrade the performance of aspect extraction. 
We also notice that Li-unified-R outperforms our full model on 14res and 14lap. 
Another insight is that the performance of RINANTE and CMLA reduced a lot in the unified tag setting, comparing with their original setting. We believe this is due to the lack of specific design to utilize sentiment information. Thus, for reference, we evaluate them under their original setting, i.e. only considering the target boundary and ignoring the sentiment polarity. The results shown in the last two rows increase drastically compared with those in the unified tag setting.

Table~\ref{tab:stage1_opt} illustrates the stage one performances of opinion term extraction. In terms of F score, our core model has again achieved the best performance compared with all existing baselines. Li-unified-R is generally not as good as our model on the restaurant datasets, but still performs very competitive and event better than our model on 14lap.
Our--TG variant model has outperformed all baselines in the laptop domain. RINANTE, CMLA and IOG only learned the mutual influence of aspect and opinion term. Compared with these baseline models, our model learns the multi-lateral information flow among the three tasks, i.e., aspect extraction, sentiment classification and opinion term extraction. In the case of opinion term extraction, it would be relatively straightforward to locate opinion terms if their sentiment polarities are given. Specifically, the $h^{\mathcal{OPT}}$ is used for unified tag prediction and thus the sentiment classification signals are backpropagated to $BLSTM^{OPT}$, therefore, the opinion prediction can leverage such information.

\begin{table*}[t]
\caption{Case study on final output. (False positives were marked with cross.)}
\label{tab:casestudy}
\resizebox{2.1\columnwidth}{!}{

\begin{tabular}{l|l|l|l|l|l}
\hline
Example       & Ground truth                                                                                                                  & Our model                                                                                                                                        & Li-unified-R+                                                                                                                                                                                                                                  & CMLA+                                                                                                                                                                                                                                                                                       & RINANTE+                                                                                                                                                                                                                                                                                                                                                                                                                                                                                \\ \hline
\begin{tabular}[c]{@{}l@{}}Rice is too dry , \\tuna was n't so\\ fresh either .\end{tabular} & \begin{tabular}[c]{@{}l@{}}(Rice-too dry-NEG), \\ (tuna-was n't so fresh-NEG)\end{tabular} &

\begin{tabular}[c]{@{}l@{}}(Rice-too dry-NEG), \\ (tuna-was n't so fresh-NEG), \\ (Rice-was n't so fresh-NEG)${\text{\xmark}}$, \\ (tuna-too dry-NEG)${\text{\xmark}}$\end{tabular} &

\begin{tabular}[c]{@{}l@{}}(Rice-dry-POS)${\text{\xmark}}$, \\ (Rice-n't-POS)${\text{\xmark}}$, \\ (tuna-dry-POS)${\text{\xmark}}$, \\
(tuna-fresh-POS)${\text{\xmark}}$\end{tabular} &

\begin{tabular}[c]{@{}l@{}}(Rice-dry-POS)${\text{\xmark}}$,  \\ (tuna-dry-POS)${\text{\xmark}}$\end{tabular} &

\begin{tabular}[c]{@{}l@{}}(tuna-dry-POS)${\text{\xmark}}$, \\ (tuna-n't so fresh either-POS)${\text{\xmark}}$\end{tabular}                                                                                                                                                                                                                                          \\ \hline
\begin{tabular}[c]{@{}l@{}}I am pleased with\\ the fast log on, speedy \\WiFi connection and \\the long battery life.\end{tabular} & \begin{tabular}[c]{@{}l@{}} (log on-pleased-POS), \\ (log on-fast-POS),\\ (WiFi connection-speedy-POS),\\ (battery life-long-POS)\end{tabular}& \begin{tabular}[c]{@{}l@{}}(log-pleased-POS)${\text{\xmark}}$,\\ (log-fast-POS)${\text{\xmark}}$,\\ (WiFi connection-speedy-POS),\\ (battery life-long-POS)\end{tabular}           & \begin{tabular}[c]{@{}l@{}}(WiFi connection-speedy-POS),\\ (battery life-long-POS)\end{tabular}                                                                                                                                               & \begin{tabular}[c]{@{}l@{}}(WiFi connection-speedy-POS),\\ (WiFi connection-long-POS)${\text{\xmark}}$, \\ (battery life-fast-POS),\\ 
(battery life-long-POS)\end{tabular}                                & \begin{tabular}[c]{@{}l@{}}(fast log-pleased-POS)${\text{\xmark}}$, \\(fast log-speedy-POS)${\text{\xmark}}$, \\ 
(WIFI-long-POS)${\text{\xmark}}$,\\ 
(battery life-long-POS)\end{tabular} \\ \hline

\begin{tabular}[c]{@{}l@{}}The service was exceptional \\- sometime there was a feeling \\that we were served by the \\army of friendly waiters .\end{tabular}

& \begin{tabular}[c]{@{}l@{}} (service-exceptional-POS), \\ (waiters-friendly-POS)\end{tabular}&

\begin{tabular}[c]{@{}l@{}}(service-exceptional-POS), \\ (waiters-friendly-POS)\end{tabular}           &

\begin{tabular}[c]{@{}l@{}}(service-exceptional-POS), \\ (waiters-friendly-POS)\\(service-feeling-POS)${\text{\xmark}}$\end{tabular}                                                                                                                                               &

\begin{tabular}[c]{@{}l@{}}
(service-exceptional-POS),\\
(waiters-friendly-POS)
\end{tabular}

& \begin{tabular}[c]{@{}l@{}}

Empty 

\end{tabular} \\ \hline

\end{tabular}

}
\end{table*}

\subsubsection{Stage two.}
After obtaining all the possible candidate triplets from the stage one, each triplet is sent to a binary classifier. The classifier was trained on the ground truth aspect and opinion pairs in the training set. 
The model performing the best on the validation set was used as the stage two classifier for evaluating both our model and baselines. (The performance of the classifier on the validation set is shown in the first row of Table~\ref{tab:stage2}). 
The last section in Table~\ref{tab:stage2} shows the performance for the final triplet extraction. We can observe that our model has achieved steady advantage over other baselines. 
In addition to evaluating the triplet, we also examine the pure pairing performance for coupling aspects and opinion terms. The results are shown in the middle section of Table~\ref{tab:stage2}. In both sections, Li-unified-R+, a variant of Li-unified implemented by us, achieved competitive performance on the first three datasets, and even slightly better than ours on 15res in the pairing evaluation.

\subsubsection{Ablation test.}
To evaluate the rationality of our model design, we also conducted ablation tests by introducing three model variants, our--$\text{BLSTM}^{\mathcal{OPT}}$, our--T and our--TG, where `--' means without the component followed behind. As we introduced before, $\text{BLSTM}^{\mathcal{OPT}}$ is expected to encode sentence contextual information which is beneficial to both unified tag prediction and opinion term extraction. 
From Table~\ref{tab:stage1_ts} and~\ref{tab:stage1_opt}, for most datasets, we can find the apparent performance reduction after removing the $\text{BLSTM}^{\mathcal{OPT}}$ module, which validates the effectiveness of this component.
Nevertheless, the contribution of TG is more complex. In the unified tag prediction task, the removal of TG module brings down the performance in all datasets reasonably. 
In the opinion term extraction, the removal even boosts the performances on 14res and 14lap datasets, especially the latter. Since TG module studies the mutual influence between aspects and opinion terms, we suspect that their mutual relation is not that strong.
Instead of bringing useful information, TG module could potentially brings in noise as well. Our assumption is validated by the classifier performance trained on gold labels in Table~\ref{tab:stage2}. 14lap and 14res have lower performances than the other two, particularly 14lap, which indicates that their target-opinion pairs are intrinsically more heterogeneous.

\subsection{Case Study}
Some triplet prediction cases are given in Table~\ref{tab:casestudy}. In general, our model outputs more reasonable results. For the first case, our model can predict more accurate opinions and sentiment polarity such as ``was n't so fresh''. However, it faces some problem in pairing prediction. The baselines cannot well capture the negated opinion. For the second case, all pipelines are hindered by the target ``log on'', our model can predict a partial target ``log''. For the third case, Li-unified-R predicts three opinions, but ``feeding'' is a wrong one, while RINANTE fails to predict tags and thus cannot output any triplet. 
One might notice that in the table, some aspects extracted in the stage one are coupled with multiple opinions, which usually brings in false positive triplets in the stage two. It might be plausible to set a heuristic rule to constrain that the pairing algorithm can only output a certain number (say equal to the number of extracted aspects) of triplets according to the classification probability. However, we did try it and found it was not consistently profitable. 

\section{Conclusions}\label{sec:conclusion}
We introduce a sentiment triplet extraction task that answers what is the aspect, how is its sentiment and why is the sentiment in one shot by coupling together aspect extraction, aspect term sentiment classification and opinion term extraction in a two-stage framework. The first stage generates candidate aspects with sentiment polarities and candidate opinion terms by utilizing mutual influence between aspects and opinion terms. The second stage pairs up the correct aspects and opinion terms. Experiments validate the feasibility and effectiveness of our model, and set a benchmark performance for this task. 

\begin{small}

\bibliographystyle{aaai}
\bibliography{aaai20}
\end{small}
\end{document}